\pdfoutput=1

\documentclass[11pt,a4paper]{article}
\usepackage{acl2015}
\usepackage{times}
\usepackage{url}
\usepackage{latexsym}

\usepackage{amsmath}
\usepackage{graphicx}
\usepackage{algorithmic}
\usepackage{algorithm}
\usepackage{enumitem}

\title{Bidirectional LSTM-CRF Models for Sequence Tagging}

\author{Zhiheng Huang \\
  Baidu research \\
  {\tt huangzhiheng@baidu.com} \\\And
  Wei Xu \\
  Baidu research \\
  {\tt xuwei06@baidu.com}  \\\And
  Kai Yu \\
  Baidu research \\
  {\tt yukai@baidu.com} 
  \\}

\date{}

\begin{document}
\renewcommand\baselinestretch{0.90}
\baselineskip=0.98\normalbaselineskip

\maketitle
\begin{abstract}
In this paper, we propose a variety of  Long Short-Term Memory (LSTM) based models for sequence tagging. These models include LSTM networks, bidirectional LSTM (BI-LSTM) networks, LSTM with a Conditional Random Field (CRF) layer (LSTM-CRF) and bidirectional LSTM with a CRF layer (BI-LSTM-CRF).  Our work is the first to apply a bidirectional LSTM CRF (denoted as BI-LSTM-CRF) model to NLP benchmark sequence tagging data sets.  We show that the BI-LSTM-CRF model can efficiently use both past and future input features thanks to a bidirectional LSTM component. It can also use sentence level tag information thanks to a CRF layer. The BI-LSTM-CRF model can produce state of the art (or close to) accuracy on POS, chunking and NER data sets. In addition, it is robust and has less dependence on word embedding as compared to previous observations.
\end{abstract}

\section{Introduction}

Sequence tagging including part of speech tagging (POS), chunking, and named entity recognition (NER) has been a classic NLP task.  It has drawn research attention for a few decades. The output of taggers can be used for down streaming applications. For example, a named entity recognizer trained on user search queries can be utilized to identify which spans of text are products, thus triggering certain products ads. Another example is that such tag information can be used by a search engine to find relevant webpages. 

Most existing sequence tagging models are linear statistical models which include Hidden Markov Models (HMM), Maximum entropy Markov models (MEMMs) \cite{mcCallum1}, and Conditional Random Fields (CRF) \cite{lafferty1}. Convolutional network based models \cite{collobert1} have been recently proposed to tackle sequence tagging problem. We denote such a model as \textit{Conv-CRF}  as it consists of a convolutional network and a CRF layer on the output (the term of \textit{sentence level log-likelihood (SSL)} was used in the original paper). The Conv-CRF model has generated promising results on sequence tagging tasks. In speech language understanding community, recurrent neural network \cite{mesnil1,yao1} and convolutional nets \cite{xu1} based models have been recently proposed. Other relevant work includes \cite{graves1,graves3} which proposed a bidirectional recurrent neural network for speech recognition.

In this paper, we propose a variety of neural network based models to sequence tagging task. These models include LSTM networks, bidirectional LSTM networks (BI-LSTM), LSTM networks with a CRF layer (LSTM-CRF), and bidirectional LSTM networks with a CRF layer (BI-LSTM-CRF). Our contributions can be summarized as follows. 1) We systematically compare the performance of aforementioned models on NLP tagging data sets; 2) Our work is the first to apply a bidirectional LSTM CRF (denoted as BI-LSTM-CRF) model to NLP benchmark sequence tagging data sets. This model can use both past and future input features thanks to a bidirectional LSTM component. In addition, this model can use sentence level tag information thanks to a CRF layer. Our model can produce state of the art (or close to) accuracy on POS, chunking and NER data sets; 3) We show that BI-LSTM-CRF model is robust and it has less dependence on word embedding as compared to previous observations \cite{collobert1}. It can produce accurate tagging performance without resorting to word embedding. 

The remainder of the paper is organized as follows. Section \ref{sec:model} describes sequence tagging models used in this paper. Section \ref{sec:train} shows the training procedure. Section \ref{sec:experiment} reports the experiments results. Section \ref{sec:discussion} discusses related research. Finally Section \ref{sec:conclusion} draws conclusions.

\section{Models} \label{sec:model}
In this section, we describe the models used in this paper: LSTM, BI-LSTM, CRF, LSTM-CRF and BI-LSTM-CRF.  

\subsection{LSTM Networks}
Recurrent neural networks (RNN) have been employed to produce promising results on a variety of tasks including language model \cite{mikolov1,mikolov2} and speech recognition \cite{graves1}.  A RNN maintains a memory based on history information, which enables the model to predict the current output conditioned on long distance features. 

\begin{figure}[htb]
	\centering
		\includegraphics[height=4cm]{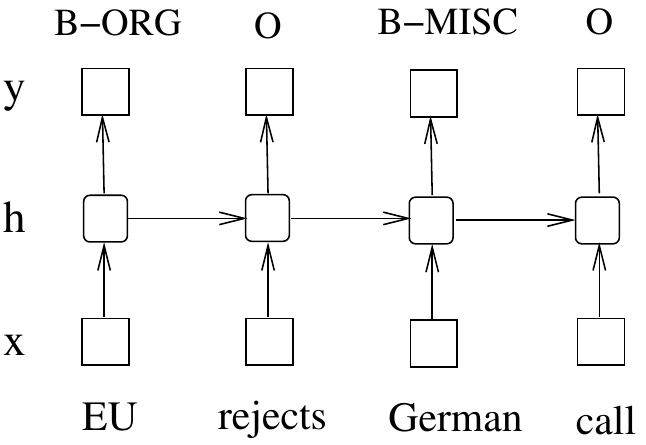}
	\caption{A simple RNN model.}
	\label{fig:rnn}
\end{figure}

Figure \ref{fig:rnn} shows the RNN structure \cite{elman1} which has an input layer $x$, hidden layer $h$ and output layer $y$. In named entity tagging context, $x$ represents input features and $y$ represents tags. Figure \ref{fig:rnn} illustrates a named entity recognition system in which each word is tagged with \textit{other (O)} or one of four entity types: \textit{Person (PER)}, \textit{Location (LOC)}, \textit{Organization (ORG)}, and \textit{Miscellaneous (MISC)}.  The sentence of \texttt{EU rejects German call to boycott British lamb . } is tagged as \texttt{B-ORG O B-MISC O O O B-MISC O O}, where \textit{B-}, \textit{I-} tags indicate beginning and intermediate positions of entities. 

An input layer represents features at time $t$. They could be one-hot-encoding for word feature, dense vector features, or sparse features. An input layer has the same dimensionality as feature size. An output layer represents a probability distribution over labels at time $t$. It has the same dimensionality as size of labels. Compared to feedforward network, a RNN introduces the connection between the previous hidden state and current hidden state (and thus the recurrent layer weight parameters).  This recurrent layer is designed to store history information. The values in the hidden and output layers are computed as follows:
\begin{eqnarray}
\textbf{h} (t) &=& f(\textbf{U}\textbf{x}(t) + \textbf{W}\textbf{h}(t-1)), \label{eq:01} \\
\textbf{y} (t) &=& g(\textbf{V}\textbf{h}(t)), \label{eq:12-2}
\end{eqnarray}
where $\textbf{U}$, $\textbf{W}$, and $\textbf{V}$ are the connection weights to be computed in training time, and 
$f(z)$ and $g(z)$ are sigmoid and softmax activation functions as follows.
\begin{eqnarray}
f(z) &=& \frac{1}{1+e^{-z}}, \\
g(z_m) &=& \frac{e^{z_m}}{\sum_k e^{z_k}}. \label{eq:softmax}
\end{eqnarray}

\begin{figure}[htb]
	\centering
		\includegraphics[height=5.5cm]{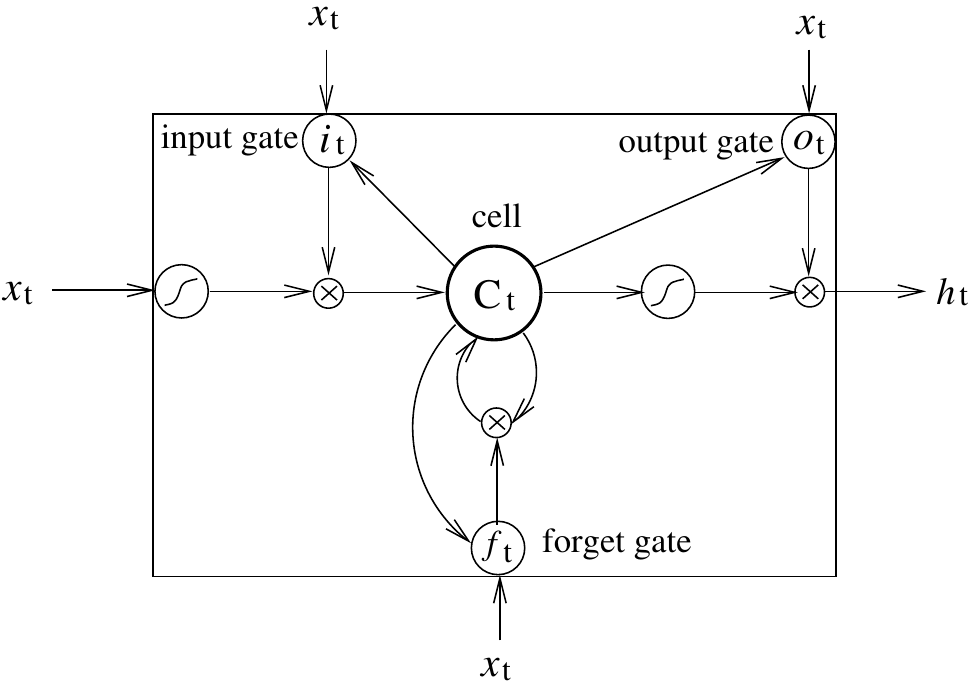}
	\caption{A Long Short-Term Memory Cell.}
	\label{fig:lstmCell}
\end{figure}

In this paper, we apply Long Short-Term Memory \cite{hochreiter1,graves1} to sequence tagging. Long Short-Term Memory networks are the same as RNNs, except that the hidden layer updates are replaced by purpose-built memory cells. As a result, they may be better at finding and exploiting long range dependencies in the data. Fig. \ref{fig:lstmCell} illustrates a single LSTM memory cell \cite{graves1}. 
The LSTM memory cell is implemented as the following:
\begin{eqnarray*}
i_t &=& \sigma (W_{xi} x_t + W_{hi} h_{t-1} + W_{ci} c_{t-1} + b_i) \\
f_t &=& \sigma (W_{xf} x_t + W_{hf} h_{t-1} + W_{cf} c_{t-1} + b_f) \\
c_t &= &f_t c_{t-1} + i_t tanh (W_{xc} x_t + W_{hc} h_{t-1} + b_c) \\
o_t &=& \sigma (W_{xo} x_t + W_{ho} h_{t-1} + W_{co} c_t + b_o) \\
h_t &=& o_t tanh (c_t)
\end{eqnarray*}
where $\sigma$ is the logistic sigmoid function, and $i$, $f$, $o$ and $c$ are the input gate, forget gate, output gate and cell vectors, all of which are the same size as the hidden vector $h$. The weight matrix subscripts have the meaning as the name suggests. For example, $W_{hi}$ is the hidden-input gate matrix, $W_{xo}$ is the input-output gate matrix etc. The weight matrices from the cell to gate vectors (e.g. $W_{ci}$) are diagonal, so element $m$ in each gate vector only receives input from element $m$ of the cell vector. 

\begin{figure}[!htb]
	\centering
		\includegraphics[width=7.5cm]{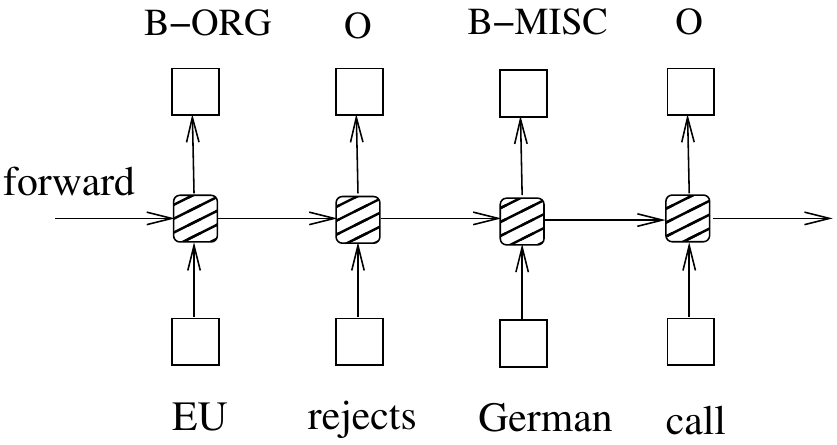}
	\caption{A LSTM network.}
	\label{fig:lstm}
\end{figure}

Fig. \ref{fig:lstm} shows a LSTM sequence tagging model which employs aforementioned LSTM memory cells (dashed boxes with rounded  corners). 

\subsection{Bidirectional LSTM Networks}

\begin{figure}[!htb]
	\centering
		\includegraphics[width=7.5cm]{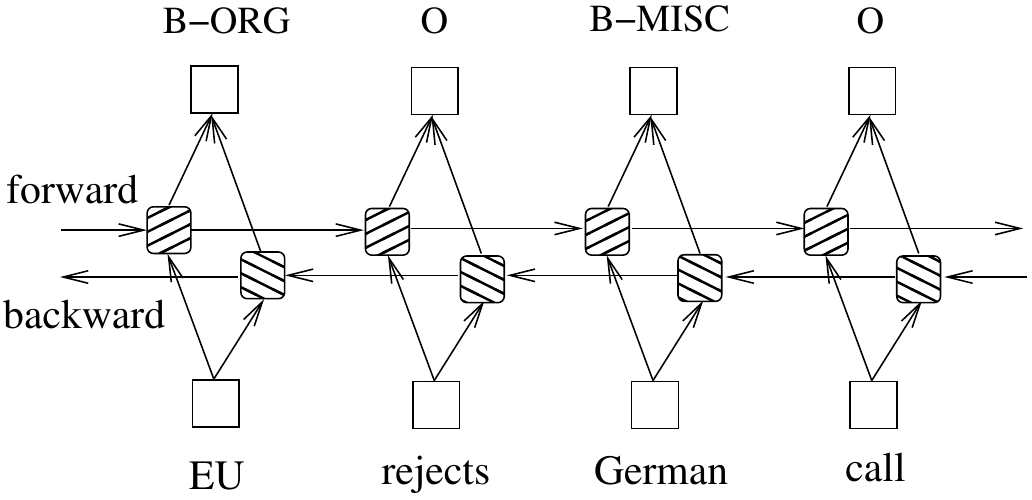}
	\caption{A bidirectional LSTM network.}
	\label{fig:biLstm}
\end{figure}

In sequence tagging task, we have access to both past and future input features for a given time, we can thus utilize a bidirectional LSTM network (Figure \ref{fig:biLstm}) as proposed in \cite{graves3}.  In doing so, we can efficiently make use of past features (via forward states) and future features (via backward states) for a specific time frame. We train bidirectional LSTM networks using back-propagation through time (BPTT)\cite{boden1}. The forward and backward passes over the unfolded network over time are carried out in a similar way to regular network forward and backward passes, except that we need to unfold the hidden states for all time steps.  We also need a special treatment at the beginning and the end of the data points. In our implementation, we do forward and backward for whole sentences and we only need to reset the hidden states to 0 at the begging of each sentence. We have batch implementation which enables multiple sentences to be processed at the same time.  

\subsection{CRF networks}
\begin{figure}[htb]
	\centering
		\includegraphics[width=6cm]{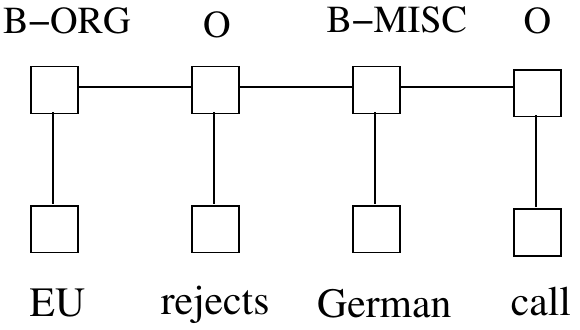}
	\caption{A CRF network.}
	\label{fig:crf}
\end{figure}

There are two different ways to make use of neighbor tag information in predicting current tags.  The first is to predict a distribution of tags for each time step and then use beam-like decoding to find optimal tag sequences. The work of maximum entropy classifier \cite{ratnaparkhi1} and Maximum entropy Markov models (MEMMs) \cite{mcCallum1} fall in this category. The second one is to focus on sentence level instead of individual positions, thus leading to Conditional Random Fields (CRF) models \cite{lafferty1} (Fig. \ref{fig:crf}). Note that the inputs and outputs are directly connected, as opposed to LSTM and  bidirectional LSTM networks where memory cells/recurrent components are employed.

It has been shown that CRFs can produce higher tagging accuracy in general. It is interesting that the relation between these two ways of using tag information bears resemblance to two ways of using input features (see aforementioned LSTM and BI-LSTM networks), and the results in this paper confirms the superiority of BI-LSTM compared to LSTM.

\subsection{LSTM-CRF networks}
\begin{figure}[!htb]
	\centering
		\includegraphics[width=7.5cm]{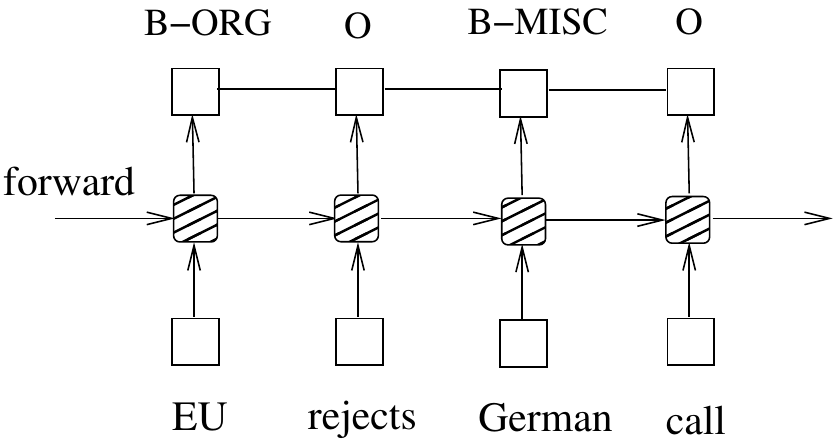}
	\caption{A LSTM-CRF model.}
	\label{fig:lstmCRF}
\end{figure}

We combine a LSTM network and a CRF network to form a LSTM-CRF model, which is shown in Fig. \ref{fig:lstmCRF}. This network can efficiently use past input features via a LSTM layer and sentence level tag information via a CRF layer. A CRF layer is represented by lines which connect consecutive output layers. A CRF layer has a state transition matrix as parameters. With such a layer, we can efficiently use past and future tags to predict the current tag, which is similar to the use of past and future input features via a bidirectional LSTM network.  We consider the matrix of scores $f_\theta([x]_1^T)$ are output by the network. We drop the input $[x]_1^T$ for notation simplification. The element $[f_\theta]_{i,t}$ of the matrix is the score output by the network with parameters $\theta$, for the sentence $[x]_1^T$ and for the $i$-th tag, at the $t$-th word. We introduce a transition score $[A]_{i,j}$ to model the transition from $i$-th state to $j$-th for a pair of consecutive time steps. Note that this transition matrix is position independent.  We now denote the new parameters for our network as $\tilde{\theta} = \theta \cup \{ [A]_{i,j} \forall i, j\}$. The score of a sentence $[x]_1^T$ along with a path of tags $[i]_1^T$ is then given by the sum of transition scores and network scores:
\begin{equation}
s([x]_1^T, [i]_1^T, \tilde{\theta}) = \sum_{t=1}^T ([A]_{[i]_{t-1}, [i]_t} + [f_{\theta}]_{[i]_t, t}).
\label{eq:score}
\end{equation}
The dynamic programming \cite{rabiner1} can be used efficiently to compute $[A]_{i,j}$ and optimal tag sequences for inference. See \cite{lafferty1} for details.

\subsection{BI-LSTM-CRF networks}
\begin{figure}[!htb]
	\centering
		\includegraphics[width=7.5cm]{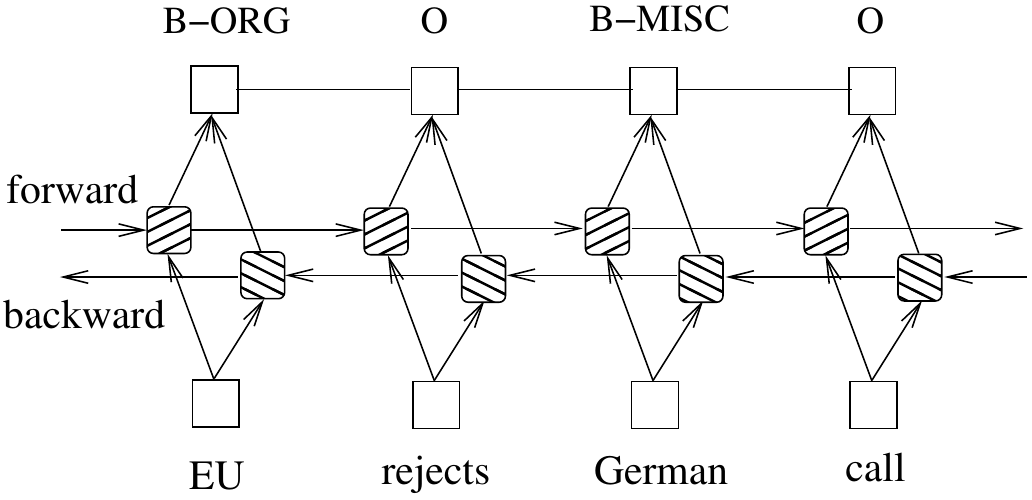}
	\caption{A BI-LSTM-CRF model.}
	\label{fig:biLstmCRF}
\end{figure}

Similar to a LSTM-CRF network, we combine a bidirectional LSTM network and a CRF network to form a BI-LSTM-CRF network (Fig. \ref{fig:biLstmCRF}). In addition to the past input features and sentence level tag information used in a LSTM-CRF model, a BI-LSTM-CRF model can use the future input features. The extra features can boost tagging accuracy as we will show in experiments.

\section{Training procedure} \label{sec:train}
All models used in this paper share a generic SGD forward and backward training procedure. We choose the most complicated model, BI-LSTM-CRF, to illustrate the training algorithm as shown in Algorithm \ref{algo:trainProcedure}.  In each epoch, we divide the whole training data to batches and process one batch at a time. Each batch contains a list of sentences which is determined by the parameter of \textit{batch size}. In our experiments, we use batch size of 100 which means to include sentences whose total length is no greater than 100. For each batch, we first run bidirectional LSTM-CRF model forward pass which includes the forward pass for both forward state and backward state of LSTM. As a result, we get the the output score $f_\theta([x]_1^T)$ for all tags at all positions. We then run CRF layer forward and backward pass to compute gradients for network output and state transition edges. After that, we can back propagate the errors from the output to the input, which includes the backward pass for both forward and backward states of LSTM. Finally we update the network parameters which include the state transition matrix $[A]_{i,j} \forall i, j$, and the original bidirectional LSTM parameters $\theta$. 
\begin{algorithm}[!hbt]
\small
\begin{algorithmic}[1]
 \FOR{each epoch}
   \FOR{each batch}
     \STATE 1) bidirectional LSTM-CRF model forward pass:
       \STATE\hspace{\algorithmicindent} forward pass for forward state LSTM
       \STATE\hspace{\algorithmicindent} forward pass for backward state LSTM
    \STATE 2) CRF layer forward and backward pass
    \STATE 3) bidirectional LSTM-CRF model backward pass:
       \STATE\hspace{\algorithmicindent} backward pass for forward state LSTM
       \STATE\hspace{\algorithmicindent} backward pass for backward state LSTM
    \STATE 4) update parameters
 \ENDFOR  
\ENDFOR 
\end{algorithmic}
\caption{Bidirectional LSTM CRF model training procedure}\label{algo:trainProcedure}
\end{algorithm}

\section{Experiments} \label{sec:experiment}
\subsection{Data}
We test LSTM, BI-LSTM, CRF, LSTM-CRF, and BI-LSTM-CRF models on three NLP tagging tasks: Penn TreeBank (PTB) POS tagging, CoNLL 2000 chunking, and CoNLL 2003 named entity tagging. Table \ref{tab:datasets} shows the size of sentences, tokens, and labels for training, validation and test sets  respectively.

POS assigns each word with a unique tag that indicates its syntactic role. In chunking, each word is tagged with its phrase type. For example, tag \textit{B-NP} indicates a word starting a noun phrase. In NER task, each word is tagged with \textit{other} or one of four entity types: \textit{Person}, \textit{Location}, \textit{Organization}, or \textit{Miscellaneous}. We use the \texttt{BIO2} annotation standard for chunking and NER tasks.  

\begin{table*}[!hbt]
\begin{center} 
\caption{Size of sentences, tokens, and labels for training, validation and test sets.}
\label{tab:datasets}
\begin{tabular}{l|c|c|c|c}
\hline
& & POS & CoNLL2000 & CoNLL2003 \\ \hline
training & sentence \# & 39831 & 8936 & 14987 \\
             & token \# & 950011 & 211727 & 204567 \\ \hline
validation & sentence \# & 1699 & N/A & 3466 \\
             & token \# & 40068 & N/A & 51578 \\ \hline
test       & sentences \# & 2415 & 2012 & 3684 \\
            & token \# & 56671 & 47377 & 46666 \\ \hline
            & label \# & 45 & 22 & 9 \\ \hline
\end{tabular}
\end{center}
\end{table*}

\subsection{Features} \label{sec:features}
We extract the same types of features for three data sets. The features can be grouped as spelling features and context features. As a result, we have 401K, 76K, and 341K features extracted for POS,  chunking and  NER data sets respectively. These features are similar to the features extracted from Stanford NER tool \cite{finkel1,wang1}. Note that we did not use extra data for POS and chunking tasks, with the exception of using Senna embedding (see Section \ref{sec:sennaEmbedding}). For NER task, we report performance with spelling and context features,  and also incrementally with Senna embedding and Gazetteer features\footnote{Downloaded from http://ronan.collobert.com/senna/}.

\subsubsection{Spelling features}
We extract the following features for a given word in addition to the lower case word features.
\begin{itemize}[noitemsep,nolistsep]
\item whether start with a capital letter
\item whether has all capital letters
\item whether has all lower case letters
\item whether has non initial capital letters
\item whether mix with letters and digits 
\item whether has punctuation 
\item letter prefixes and suffixes (with window size of 2 to 5)
\item whether has apostrophe end ('s)
\item letters only, for example, I. B. M. to IBM
\item non-letters only, for example, A. T. \&T. to ..\&
\item word pattern feature, with capital letters, lower case letters, and digits mapped to `A', `a' and `0' respectively, for example, D56y-3 to A00a-0
\item word pattern summarization feature, similar to word pattern feature but with consecutive identical characters removed. For example, D56y-3 to A0a-0
\end{itemize}

\subsubsection{Context features}
For word features in three data sets, we use uni-gram features and bi-grams features. For POS features in CoNLL2000 data set and POS \& CHUNK features in CoNLL2003 data set, we use unigram, bi-gram and tri-gram features. 

\subsubsection{Word embedding} \label{sec:sennaEmbedding}
It has been shown in \cite{collobert1} that word embedding plays a vital role to improve sequence tagging performance. We downloaded\footnote{http://ronan.collobert.com/senna/} the embedding which has 130K vocabulary size and each word corresponds to a 50-dimensional embedding vector. To use this embedding, we simply replace the one hot encoding word representation with its corresponding 50-dimensional vector.

\subsubsection{Features connection tricks}
\begin{figure}[!htb]
	\centering
		\includegraphics[width=8cm]{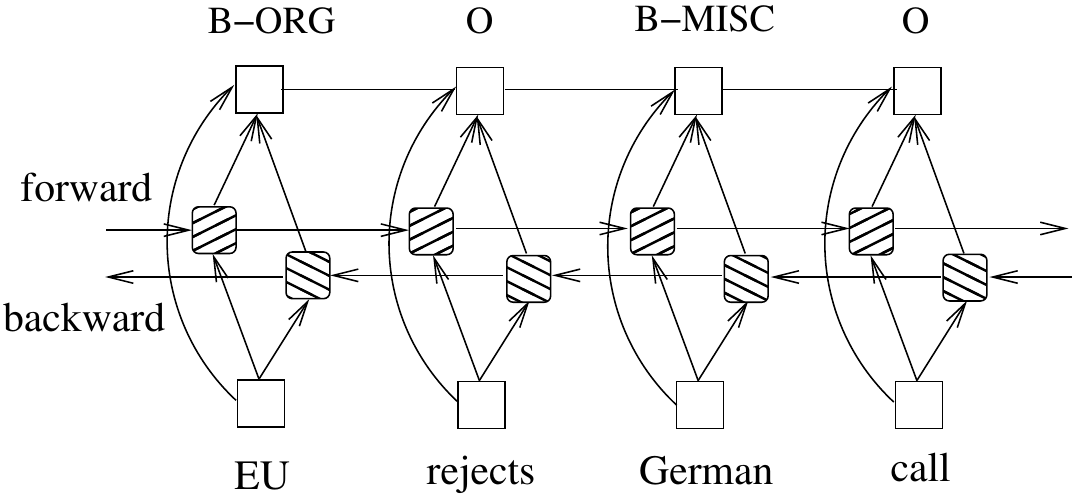}
	\caption{A BI-LSTM-CRF model with MaxEnt features.}
	\label{fig:biLstmCRFMaxEnt}
\end{figure}

We can treat spelling and context features the same as word features. That is, the inputs of networks include both word, spelling and context features. However, we find that direct connections from spelling and context features to outputs accelerate training and they result in very similar tagging accuracy. Fig. \ref{fig:biLstmCRFMaxEnt} illustrates this network in which features have direct connections to outputs of networks. We will report all tagging accuracy using this connection. We note that this usage of features has the same flavor of Maximum Entropy features as used in \cite{mikolov2}. The difference is that features collision may occur in \cite{mikolov2} as feature hashing technique has been adopted. Since the output labels in sequence tagging data sets are less than that of language model (usually hundreds of thousands), we can afford to have full connections between features and outputs to avoid  potential feature collisions.

\subsection{Results}
We train LSTM, BI-LSTM, CRF, LSTM-CRF and BI-LSTM-CRF models for each data set. We have two ways to initialize word embedding: \textit{Random} and \textit{Senna}. We randomly initialize the word embedding vectors in the first category, and use Senna word embedding in the second category. For each category, we use identical feature sets,  thus different results are solely due to different networks. We train models using training data and monitor performance on validation data. As chunking data do not have a validation data set, we use part of  training data for validation purpose.  

We use a learning rate of 0.1 to train models. We set hidden layer size to 300 and found that model performance is not sensitive to hidden layer sizes. The training for three tasks require less than 10 epochs to converge and it in general takes less than a few hours. We report models' performance on  test datasets in Table \ref{tab:results}, which also lists the best results in \cite{collobert1}, denoted as Conv-CRF. The POS task is evaluated by computing per-word accuracy, while the chunk and NER tasks are evaluated by computing F1 scores over chunks. 

\subsubsection{Comparison with Cov-CRF networks}
We have three baselines: LSTM, BI-LSTM and CRF. LSTM is the weakest baseline for all three data sets. The BI-LSTM performs close to CRF on POS and chunking datasets, but is worse than CRF on NER data set. The CRF forms strong baselines in our experiments. For random category,  CRF models outperform Conv-CRF models for all three data sets. For Senna category, CRFs outperform Conv-CRF for POS task, while underperform for chunking and NER task. LSTM-CRF models outperform CRF models for all data sets in both random and Senna categories. This shows the effectiveness of the forward state LSTM component in modeling sequence data. The BI-LSTM-CRF models further improve LSTM-CRF models and they lead to the best tagging performance for all cases except for POS data at random category, in which LSTM-CRF model is the winner. The numbers in parentheses for CoNLL 2003 under Senna categories are generated with Gazetteer features. 

It is interesting that our best model BI-LSTM-CRF has less dependence on Senna word embedding compared to Conv-CRF model. For example, the tagging difference between BI-LSTM-CRF model for random and Senna categories are 0.12\%, 0.33\%, and 4.57\% for POS, chunking and NER data sets respectively. In contrast, the Conv-CRF model heavily relies on Senna embedding to get good tagging accuracy. It has the tagging difference of 0.92\%, 3.99\% and 7.20\% between random and Senna category for POS, chunking and NER data sets respectively.

\begin{table*}[!hbt]
\begin{center} 
\caption{Comparison of tagging performance on POS, chunking and NER tasks for various models.}
\label{tab:results}
\begin{tabular}{l|l|c|c|c}
\hline
& &  POS & CoNLL2000 & CoNLL2003 \\ \hline
& Conv-CRF  \cite{collobert1} & 96.37 & 90.33 &  81.47 \\ 
& LSTM & 97.10 & 92.88 & 79.82 \\
& BI-LSTM & 97.30 & 93.64 & 81.11 \\
Random & CRF & 97.30 & 93.69 & 83.02 \\ 
& LSTM-CRF & \textbf{97.45} & 93.80 & 84.10 \\
& BI-LSTM-CRF & 97.43 & \textbf{94.13} & \textbf{84.26} \\ \hline
& Conv-CRF \cite{collobert1} & 97.29 & 94.32 &  88.67 (89.59) \\ 
& LSTM & 97.29 & 92.99 & 83.74 \\
& BI-LSTM & 97.40 & 93.92 & 85.17 \\
Senna & CRF &  97.45 & 93.83 & 86.13 \\ 
 & LSTM-CRF & 97.54 & 94.27 & 88.36 \\
& BI-LSTM-CRF & \textbf{97.55} & \textbf{94.46} & \textbf{88.83} (\textbf{90.10}) \\ \hline
\end{tabular}
\end{center}
\end{table*}

\subsubsection{Model robustness}
To estimate the robustness of models with respect to engineered features (spelling and context features), we train LSTM, BI-LSTM, CRF, LSTM-CRF, and BI-LSTM-CRF models with word features only (spelling and context features removed). Table \ref{tab:results2} shows tagging performance of proposed models for POS, chunking, and NER data sets using Senna word embedding. The numbers in parentheses indicate the performance degradation compared to the same models but using spelling and context features. CRF models' performance is significantly degraded with the removal of spelling and context features. This reveals the fact that CRF models heavily rely on engineered features to obtain good performance. On the other hand, LSTM based models, especially BI-LSTM and BI-LSTM-CRF models are more robust and they are less affected by the removal of engineering features.  For all three tasks, BI-LSTM-CRF models result in the highest tagging accuracy. For example, It achieves the F1 score of 94.40 for CoNLL2000 chunking, with slight degradation (0.06) compared to the same model but using spelling and context features.

\begin{table*}[!hbt]
\begin{center} 
\caption{Tagging performance on POS, chunking and NER tasks with only word features.}
\label{tab:results2}
\begin{tabular}{l|l|c|c|c}
\hline
& &  POS & CoNLL2000 & CoNLL2003 \\ \hline
& LSTM & 94.63 (-2.66) & 90.11 (-2.88) &  75.31 (-8.43)\\
& BI-LSTM & 96.04 (-1.36) & 93.80 (-0.12) &  83.52 (-1.65)\\
Senna & CRF &  94.23 (-3.22) & 85.34 (-8.49) &  77.41 (-8.72)\\ 
 & LSTM-CRF & 95.62 (-1.92) & 93.13 (-1.14) &  81.45 (-6.91)\\
& BI-LSTM-CRF & \textbf{96.11} (-1.44)& \textbf{94.40} (-0.06) & \textbf{84.74} (-4.09) \\ \hline
\end{tabular}
\end{center}
\end{table*}

\subsubsection{Comparison with existing systems}
For POS data set, we achieved state of the art tagging accuracy with or without the use of extra data resource. POS data set has been extensively tested and the past improvement can be realized in Table \ref{tab:pos}. Our test accuracy is 97.55\% which is significantly better than others in the confidence level of 95\%. In addition, our BI-LSTM-CRF model already reaches a good accuracy without the use of the Senna embedding. 

\begin{table*}[!hbt]
\begin{center} 
\caption{Comparison of tagging accuracy of different models for POS.}
\label{tab:pos}
\begin{tabular}{l|c|c}
\hline
System & accuracy & extra data \\ \hline
Maximum entropy cyclic dependency  & 97.24 & No \\
network \cite{toutanova1} & &  \\ 
SVM-based tagger \cite{gimenez1} & 97.16 & No \\ 
Bidirectional perceptron learning \cite{shen1} & 97.33 & No \\ 
Semi-supervised condensed nearest neighbor & 97.50 & Yes \\ 
\cite{soegaard1} & & \\ 
CRFs with structure regularization \cite{sun1} & 97.36 & No \\  
Conv network tagger \cite{collobert1} & 96.37 & No \\ 
Conv network tagger (senna) \cite{collobert1} & 97.29 & Yes \\ \hline
BI-LSTM-CRF (ours) & \textbf{97.43} & No  \\
BI-LSTM-CRF (Senna) (ours) &  \textbf{97.55} & Yes \\ \hline
\end{tabular}
\end{center}
\end{table*}

\begin{table*}[!hbt]
\begin{center} 
\caption{Comparison of F1 scores of different models for chunking.}
\label{tab:chunk}
\begin{tabular}{l|c}
\hline
System & accuracy  \\ \hline
SVM classifier \cite{kudo1} & 93.48  \\
SVM classifier \cite{kudo2} & 93.91  \\
Second order CRF \cite{sha1} & 94.30  \\
Specialized HMM + voting scheme \cite{shen2} & \textbf{95.23} \\
Second order CRF \cite{mcdonald1} & 94.29 \\ 
Second order CRF \cite{sun2} & 94.34 \\ 
Conv-CRF \cite{collobert1} & 90.33  \\
Conv network tagger (senna) \cite{collobert1} & 94.32 \\ \hline
BI-LSTM-CRF (ours) & 94.13  \\
BI-LSTM-CRF (Senna) (ours) &  94.46\\\hline
\end{tabular}
\end{center}
\end{table*}

\begin{table*}[!hbt]
\begin{center} 
\caption{Comparison of F1 scores of different models for NER.}
\label{tab:ner}
\begin{tabular}{l|c}
\hline
System & accuracy  \\ \hline
Combination of HMM, Maxent etc. \cite{florian1} & 88.76  \\
MaxEnt classifier \cite{chieu1} & 88.31  \\
Semi-supervised model combination \cite{ando1} & 89.31 \\
Conv-CRF \cite{collobert1} & 81.47  \\
Conv-CRF (Senna + Gazetteer) \cite{collobert1} & 89.59 \\ 
CRF with Lexicon Infused Embeddings \cite{passos1} & \textbf{90.90} \\ \hline
BI-LSTM-CRF (ours) & 84.26  \\
BI-LSTM-CRF (Senna + Gazetteer) (ours) &  90.10\\\hline
\end{tabular}
\end{center}
\end{table*}

All chunking systems performance is shown in table \ref{tab:chunk}. Kudo et al. won the CoNLL 2000 challenge with a F1 score of 93.48\%. Their approach was a SVM based classifier. They later improved the results up to 93.91\%. Recent work include the CRF based models \cite{sha1,mcdonald1,sun2}. More recent is \cite{shen2} which obtained 95.23\% accuracy with a voting classifier scheme, where each classifier is trained on different tag representations (\texttt{IOB}, \texttt{IOE}, etc.). Our model outperforms all reported systems except \cite{shen2}.

The performance of all systems for NER is shown in table \ref{tab:ner}. \cite{florian1} presented the best system at the NER CoNLL 2003 challenge, with 88.76\%
F1 score. They used a combination of various machine-learning classifiers. The second best performer of CoNLL 2003 \cite{chieu1}  was 88.31\% F1, also with the help of an external gazetteer.  Later, \cite{ando1} reached 89.31\% F1 with a semi-supervised approach. The best F1 score of 90.90\% was reported in \cite{passos1} which  employed a new form of learning word embeddings that can leverage information from relevant lexicons to improve the representations. Our model can achieve the best F1 score of 90.10 with both Senna embedding and gazetteer features. It has a lower F1 score than \cite{passos1} , which may be  due to the fact that different word embeddings were employed. With the same Senna embedding, BI-LSTM-CRF slightly outperforms Conv-CRF (90.10\% vs. 89.59\%). However, BI-LSTM-CRF significantly outperforms Conv-CRF (84.26\% vs. 81.47\%) if random embedding is used.

\section{Discussions} \label{sec:discussion}
Our work is close to the work of \cite{collobert1} as both of them utilized deep neural networks for sequence tagging. While their work used convolutional neural networks, ours used bi-directional LSTM networks. 

Our work is also close to the work of \cite{hammerton1,yao2} as all of them employed LSTM network for tagging. The performance in \cite{hammerton1} was not impressive. The work in \cite{yao2} did not make use of bidirectional LSTM and CRF layers and thus the tagging accuracy may be suffered.

Finally, our work is related to the work of \cite{wang1} which concluded that non-linear architecture offers no benefits in a high-dimensional discrete feature space. We showed that with the bi-directional LSTM CRF model, we consistently obtained better tagging accuracy than a single CRF model with identical feature sets.

\section{Conclusions} \label{sec:conclusion}
In this paper, we systematically compared the performance of LSTM networks based models for sequence tagging. We presented the first work of applying a BI-LSTM-CRF model to NLP benchmark sequence tagging data. Our model can produce state of the art (or close to) accuracy on POS, chunking and NER data sets. In addition, our model is robust and it has less dependence on word embedding as compared to the observation in \cite{collobert1}. It can achieve accurate tagging accuracy without resorting to word embedding. 


\begin{thebibliography}{}

\bibitem[\protect\citename{Ando and Zhang.}2005]{ando1}
R. K. Ando and T. Zhang.
\newblock 2005.
\newblock {\em  A framework for learning predictive structures from multiple tasks and
unlabeled data}.
\newblock Journal of Machine Learning Research (JMLR).

\bibitem[\protect\citename{Boden.}2002]{boden1}
M. Boden.
\newblock 2002.
\newblock {\em  A Guide to Recurrent Neural Networks and Back-propagation}.
\newblock In the Dallas project.


\bibitem[\protect\citename{Chieu.}2003]{chieu1}
H. L. Chieu.
\newblock 2003.
\newblock {\em  Named entity recognition with a maximum entropy approach}.
\newblock Proceedings of CoNLL.

\bibitem[\protect\citename{Collobert et al.}2011]{collobert1}
R. Collobert, J. Weston, L. Bottou, M. Karlen, K. Kavukcuoglu and P. Kuksa.
\newblock 2011.
\newblock {\em Natural Language Processing (Almost) from Scratch}.
\newblock Journal of Machine Learning Research (JMLR).

\bibitem[\protect\citename{Elman}1990]{elman1}
J. L. Elman.
\newblock 1990.
\newblock {\em Finding structure in time}.
\newblock Cognitive Science.

\bibitem[\protect\citename{Finkel et al.}2005]{finkel1}
J. R. Finkel, T. Grenager, and C. Manning.
\newblock 2005.
\newblock {\em Incorporating Non-local Information into Information Extraction Systems by Gibbs Sampling}.
\newblock Proceedings of ACL.

\bibitem[\protect\citename{Florian et al.}2003]{florian1}
R. Florian, A. Ittycheriah, H. Jing, and T. Zhang.
\newblock 2003.
\newblock {\em  Named entity recognition through classifier combination}.
\newblock Proceedings of NAACL-HLT.

\bibitem[\protect\citename{Gimenez and Marquez}2004]{gimenez1}
J. Gimenez and L. Marquez.
\newblock 2004.
\newblock {\em SVMTool: A general POS tagger generator based on support vector machines}.
\newblock Proceedings of LREC.

\bibitem[\protect\citename{Graves et al.}2005]{graves1}
A. Graves and J. Schmidhuber.
\newblock 2005.
\newblock {\em Framewise Phoneme Classification with Bidirectional LSTM and Other Neural Network Architectures}.
\newblock Neural Networks.


\bibitem[\protect\citename{Graves et al.}2013]{graves3}
A. Graves, A. Mohamed, and G. Hinton.
\newblock 2013.
\newblock {\em Speech Recognition with Deep Recurrent Neural Networks}.
\newblock arxiv.

\bibitem[\protect\citename{Hammerton}2003]{hammerton1}
J. Hammerton.
\newblock 2003.
\newblock {\em Named Entity Recognition with Long Short-Term Memory}.
\newblock Proceedings of HLT-NAACL.

\bibitem[\protect\citename{Hochreiter and Schmidhuber}1997]{hochreiter1}
S. Hochreiter and J. Schmidhuber.
\newblock 1997.
\newblock {\em Long short-term memory}.
\newblock Neural Computation, 9(8):1735-1780.

\bibitem[\protect\citename{Kudo and Matsumoto}2000]{kudo1}
T. Kudo and Y. Matsumoto.
\newblock 2000.
\newblock {\em Use of support vector learning for chunk identification}.
\newblock Proceedings of CoNLL.

\bibitem[\protect\citename{Kudo and Matsumoto}2001]{kudo2}
T. Kudo and Y. Matsumoto.
\newblock 2001.
\newblock {\em Chunking with support vector machines}.
\newblock Proceedings of NAACL-HLT.

\bibitem[\protect\citename{Lafferty et al.}2001]{lafferty1}
J. Lafferty, A. McCallum, and F. Pereira.
\newblock 2001.
\newblock {\em Conditional random fields: Probabilistic models for segmenting
and labeling sequence data}.
\newblock Proceedings of ICML.

\bibitem[\protect\citename{McCallum et al.}2000]{mcCallum1}
A. McCallum, D. Freitag, and F. Pereira. 
\newblock 2000.
\newblock {\em Maximum entropy Markov models for information extraction and segmentation}.
\newblock Proceedings of ICML.

\bibitem[\protect\citename{Mcdonald et al.}2005]{mcdonald1}
R. Mcdonald , K. Crammer , and F. Pereira.
\newblock 2005.
\newblock {\em Flexible text segmentation with structured multilabel classification}.
\newblock Proceedings of HLT-EMNLP.

\bibitem[\protect\citename{Mesnil et al.}2013]{mesnil1}
G. Mesnil, X. He, L. Deng, and Y. Bengio.
\newblock 2013.
\newblock {\em Investigation of recurrent-neural-network architectures and learning methods for language understanding}.
\newblock Proceedings of INTERSPEECH.

\bibitem[\protect\citename{Mikolov et al.}2010]{mikolov1}
T. Mikolov, M. Karafiat, L. Burget, J. Cernocky, and S. Khudanpur.
\newblock 2010.
\newblock {\em Recurrent neural network based language model}.
\newblock INTERSPEECH.

\bibitem[\protect\citename{Mikolov et al.}2011]{mikolov2}
T. Mikolov, A. Deoras, D. Povey,  L. Burget, J. Eernocky.
\newblock 2011.
\newblock {\em Strategies for Training Large Scale Neural Network Language Models}.
\newblock Proceedings of ASRU.

\bibitem[\protect\citename{Mikolov et al.}2013]{mikolov3}
T. Mikolov, I. Sutskever, K. Chen, G. Corrado, and J. Dean.
\newblock 2013.
\newblock {\em Distributed Representations of Words and Phrases and their Compositionality}.
\newblock Proceedings of NIPS.

\bibitem[\protect\citename{Passos et al.}2014]{passos1}
A. Passos, V. Kumar, and A. McCallum.
\newblock 2014.
\newblock {\em Lexicon Infused Phrase Embeddings for Named Entity Resolution}.
\newblock Proceedings of CoNLL.


\bibitem[\protect\citename{Rabiner}1989]{rabiner1}
L. R. Rabiner.
\newblock 1989.
\newblock {\em A tutorial on hidden Markov models and selected applications in speech recognition}.
\newblock Proceedings of the IEEE.

\bibitem[\protect\citename{Ratnaparkhi}1996]{ratnaparkhi1}
A. Ratnaparkhi.
\newblock 1996.
\newblock {\em A maximum entropy model for part-of-speech tagging}.
\newblock Proceedings of EMNLP.

\bibitem[\protect\citename{Sha and Pereira}2003]{sha1}
F. Sha and F. Pereira.
\newblock 2003.
\newblock {\em Shallow parsing with conditional random fields}.
\newblock Proceedings of NAACL.

\bibitem[\protect\citename{Shen et al.}2007]{shen1}
L. Shen, G. Sara, and A. K. Joshi.
\newblock 2007.
\newblock {\em Guided learning for bidirectional sequence classification}.
\newblock Proceedings of ACL.

\bibitem[\protect\citename{Shen and Sarkar}2005]{shen2}
H. Shen and A. Sarkar.
\newblock 2005.
\newblock {\em  Voting between multiple data representations for text chunking}.
\newblock Canadian AI.

\bibitem[\protect\citename{Soegaard}2011]{soegaard1}
A. Soegaard.
\newblock 2011.
\newblock {\em Semi-supervised condensed nearest neighbor for part-of-speech tagging}.
\newblock Proceedings of ACL-HLT.

\bibitem[\protect\citename{Sun}2014]{sun1}
X. Sun.
\newblock 2014.
\newblock {\em Structure Regularization for Structured Prediction}.
\newblock Proceedings of NIPS.

\bibitem[\protect\citename{Sun et al.}2008]{sun2}
X. Sun, L.P. Morency, D. OKanohara and J. Tsujii.
\newblock 2008.
\newblock {\em Modeling Latent-Dynamic in Shallow Parsing: A Latent Conditional Model with Improved Inference}.
\newblock Proceedings of COLING.

\bibitem[\protect\citename{Toutanova et al.}2003]{toutanova1}
K. Toutanova, D. Klein, C. Manning, and Y. Singer.
\newblock 2003.
\newblock {\em Feature-Rich Part-of-Speech Tagging with a Cyclic Dependency Network}.
\newblock Proceedings of HLT-NAACL.

\bibitem[\protect\citename{Wang and Manning}2013]{wang1}
M. Wang and C. D. Manning. 
\newblock 2013.
\newblock {\em Effect of Non-linear Deep Architecture in Sequence Labeling}.
\newblock Proceedings of IJCNLP.

\bibitem[\protect\citename{Xu and Sarikaya}2013]{xu1}
P. Xu and R. Sarikaya.
\newblock 2013.
\newblock {\em Convolutional neural network based triangular
CRF for joint intent detection and slot filling}.
\newblock Proceedings of ASRU.

\bibitem[\protect\citename{Yao et al.}2014]{yao1}
K. S. Yao, B. L. Peng, G. Zweig, D. Yu, X. L. Li, and F. Gao.
\newblock 2014.
\newblock {\em Recurrent conditional random fields for language understanding}.
\newblock ICASSP.

\bibitem[\protect\citename{Yao et al.}2014]{yao2}
K. S. Yao, B. Peng, Y. Zhang, D. Yu, G. Zweig, and Y. Shi.
\newblock 2014.
\newblock {\em Spoken Language Understanding using Long Short-Term Memory Neural Networks}.
\newblock IEEE SLT.

\end{thebibliography}

\end{document}